\documentclass{article}

     \usepackage[final]{neurips_2018}

\usepackage[utf8]{inputenc} 
\usepackage[T1]{fontenc}    
\usepackage{hyperref}       
\usepackage{url}            
\usepackage{booktabs}       
\usepackage{amsfonts}       
\usepackage{nicefrac}       
\usepackage{microtype}      
\usepackage{amsmath}

\title{Uncertainty Sampling is Preconditioned Stochastic Gradient Descent on Zero-One Loss}

\author{
  Stephen Mussmann \\ 
  Department of Computer Science\\
  Stanford University\\
  Stanford, CA \\
  \texttt{mussmann@stanford.edu} \\
   \And
   Percy Liang \\
   Department of Computer Science\\
   Stanford University \\
   Stanford, CA \\
   \texttt{pliang@cs.stanford.edu} \\
}

\usepackage{amsthm}
\usepackage{amsmath}
\usepackage{amssymb}
\usepackage{algorithm}
\usepackage{algorithmic}
\usepackage{mathtools}
\usepackage{caption}
\usepackage{subcaption}
\usepackage{wrapfig}

\theoremstyle{plain}
    \newtheorem{theorem}{Theorem}
    \newtheorem{lemma}[theorem]{Lemma}
    \newtheorem{definition}[theorem]{Definition}
    \newtheorem{proposition}[theorem]{Proposition}
    \newtheorem{assumption}[theorem]{Assumption}
    \newtheorem{corollary}[theorem]{Corollary}

    \newtheoremstyle{TheoremNum}
        {\topsep}{\topsep}              
        {\itshape}                      
        {}                              
        {\bfseries}                     
        {.}                             
        { }                             
        {\thmname{#1}\thmnote{ \bfseries #3}}
    \theoremstyle{TheoremNum}
    \newtheorem{thm}{Theorem}

\DeclareMathOperator*{\argmin}{arg\,min}

\usepackage{color}

\newcommand\validparams{\Theta_{\text{regular}}}
\newcommand\ellbound{M_\ell}

\newcommand\nseed{n_\text{seed}}

\newcommand\lambdamin{\lambda_{\text{min}}}

\newcommand\sD{\ensuremath{\mathcal{D}}}

\newcommand\BP{\ensuremath{\mathbb{P}}}

\newcommand\R{\ensuremath{\mathbb{R}}} 
\newcommand\eqdef{\ensuremath{\stackrel{\rm def}{=}}} 
\newcommand\refeqn[1]{(\ref{eqn:#1})}

\newcommand\refsec[1]{Section~\ref{sec:#1}}

\ifthenelse{\isundefined{\definition}}{\newtheorem{definition}{Definition}}{}
\ifthenelse{\isundefined{\assumption}}{\newtheorem{assumption}{Assumption}}{}
\ifthenelse{\isundefined{\hypothesis}}{\newtheorem{hypothesis}{Hypothesis}}{}
\ifthenelse{\isundefined{\proposition}}{\newtheorem{proposition}{Proposition}}{}
\ifthenelse{\isundefined{\theorem}}{\newtheorem{theorem}{Theorem}}{}
\ifthenelse{\isundefined{\lemma}}{\newtheorem{lemma}{Lemma}}{}
\ifthenelse{\isundefined{\corollary}}{\newtheorem{corollary}{Corollary}}{}
\ifthenelse{\isundefined{\alg}}{\newtheorem{alg}{Algorithm}}{}
\ifthenelse{\isundefined{\example}}{\newtheorem{example}{Example}}{}
\newcommand{\E}{\ensuremath{\mathbb{E}}} 

\begin{document}

\maketitle

\begin{abstract} 
Uncertainty sampling, a popular active learning algorithm, is used to reduce the amount of data required to learn a classifier, but it has been observed in practice to converge to different parameters depending on the initialization and sometimes to even better parameters than standard training on all the data. In this work, we give a theoretical explanation of this phenomenon, showing that uncertainty sampling on a convex loss can be interpreted as performing a preconditioned stochastic gradient step on a smoothed version of the population zero-one loss that converges to the population zero-one loss. Furthermore, uncertainty sampling moves in a descent direction and converges to stationary points of the smoothed population zero-one loss. Experiments on synthetic and real datasets support this connection.

\end{abstract} 

\section{Introduction}

Active learning algorithms aim to learn parameters with less data by querying
labels adaptively. However, since such algorithms change the sampling
distribution, they can introduce bias in the learned parameters. While there
has been some work to understand this \citep{schutze2006performance,bach2007active,dasgupta2008hierarchical,beygelzimer2009importance},
the most common algorithm, ``uncertainty
sampling'' \citep{lewis1994sequential,settles2010active}, remains elusive.
One of the oddities of uncertainty sampling is that sometimes the bias is \emph{helpful}:
uncertainty sampling with a subset of the data can yield lower error than random sampling on
\emph{all} the data \citep{schohn2000less,bordes2005fast,chang2017active}.
But sometimes, uncertainty sampling can vastly underperform,
and in general, different initializations can yield different parameters asymptotically.
Despite the wealth of theory on active learning \citep{balcan2006agnostic, hanneke2014theory},
a theoretical account of uncertainty sampling is lacking.

\begin{figure}[!t]
\centering
\includegraphics[width=0.99 \columnwidth]{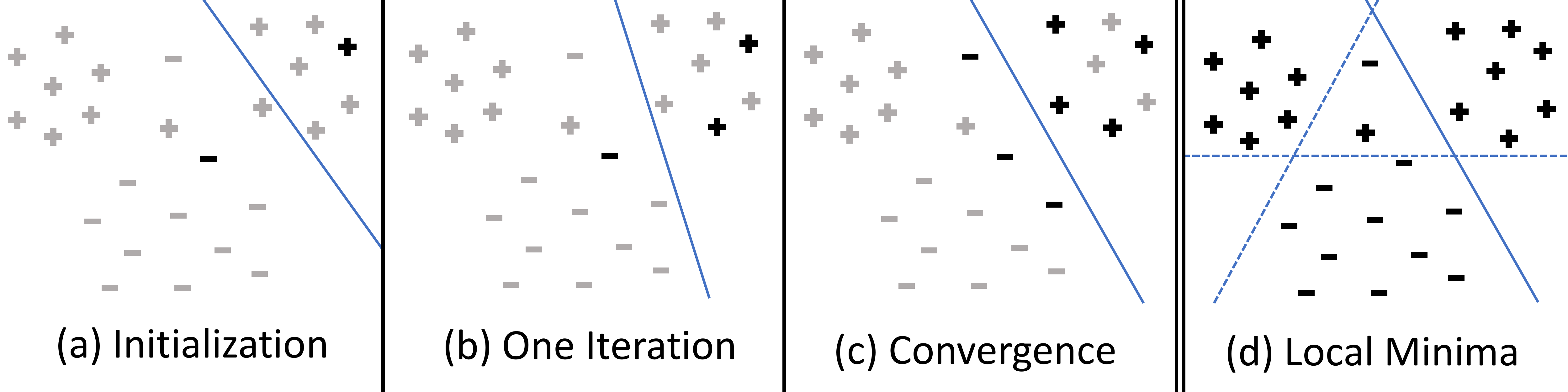}
\caption{A typical run of uncertainty sampling. Each iteration, uncertainty sampling chooses to label a point close to the current decision boundary.
  (a) Random initialization of uncertainty sampling.
  (b) A point close to the decision boundary is added and the decision boundary is updated.
  (c) Several more points close the decision boundary are added until convergence.
  We see that uncertainty sampling uses only a fraction of the data, but converges to a local minimum of the zero-one loss.
  (d) There are three different local minima of the zero-one loss,
  where the horizontal linear classifier is much more preferable than the other two.
 }
\label{fig:bad_convergence}
\end{figure}

In this paper, we characterize the dynamics of a streaming variant of
uncertainty sampling to
explain the bias introduced. We introduce a smoothed version of the zero-one loss which approximates and converges to the zero-one loss. We show that uncertainty
sampling, which minimizes a \emph{convex surrogate loss on all the points so far},
is asymptotically performing a preconditioned \emph{stochastic gradient step on the smoothed (non-convex) population zero-one loss}. Furthermore, each uncertainty sampling iterate in expectation moves in a descent direction of the smoothed population zero-one loss, unless the parameters are at an approximate stationary point. In addition, uncertainty sampling converges to a stationary point of the smoothed population zero-one loss.
This explains why uncertainty sampling sometimes achieves lower zero-one
loss than random sampling, since that is approximately the quantity it implicitly optimizes.
At the same time, as the zero-one loss is non-convex, we can get stuck in
a local minimum with higher zero-one loss (see Figure~\ref{fig:bad_convergence}).

Empirically, we validate the properties of uncertainty sampling on a
simple synthetic dataset for intuition as well as 22 real-world datasets.
Our new connection between uncertainty sampling and zero-one loss minimization clarifies the
importance of a sufficiently large seed set, rather than
using a single point per class, as
is commonly done in the literature \citep{tong2001support,yang2016benchmark}.

\section{Setup}

We focus on binary classification. Let $z=(x,y)$ be a
data point, where $x \in \R^k$ is the input and $y \in \{-1,1\}$ is the
label, drawn from some unknown true data distribution $z \sim p^*$.
Assume we have a
scoring function $S(x, \theta)$, where $\theta \in \R^d$ are the parameters;
for linear models, we have $S(x, \theta) = \theta \cdot \phi(x)$, where $\phi :
\R^k \to \R^d$ is the feature map.

Given parameters $\theta$, we predict $1$ if $S(x, \theta) > 0$ and $-1$ otherwise,
and therefore err when $y$ and $S(x, \theta)$ have opposite signs.
Define $Z(\theta)$ to be the zero-one loss (misclassification rate) over the data distribution, the evaluation metric of interest:
\begin{align}
Z(\theta) &\eqdef \E_{(x,y) \sim p^*}[H(-y S(x, \theta))],
\end{align}
where $H$ is the Heaviside step function:
\begin{align}
H(x) \eqdef
\begin{cases}
0 & x<0, \\
\frac12 & x=0, \\
1 & x>0.
\end{cases}
\end{align}

Note that the \emph{training} zero-one loss is step-wise
constant, and the gradient is $0$ almost everywhere. However, assuming the probability density function (PDF) of $p^*$ is
smooth, the \emph{population} zero-one loss is differentiable at
most parameters, a fact that will be shown later.

Since minimizing the zero-one loss is computationally intractable \citep{feldman2012agnostic},
it is common to define a convex surrogate $\ell((x,y), \theta) = \psi(y S(x,\theta))$ which upper bounds the zero-one loss;
for example, the logistic loss is given by $\psi(s) = \log (1 + e^{-s})$.
Given a labeled dataset $\sD = \{ z_1, \dots, z_n \}$, we can define the estimator that minimizes
the sum of the loss plus regularization:
\begin{align}
  \label{eqn:regerm}
  \theta_\sD \eqdef \argmin_\theta \sum_{z \in \sD} \ell(z, \theta) + \lambda \|\theta\|_2^2.
\end{align}
This can often be solved efficiently via convex optimization.

\paragraph{Passive learning: random sampling.}
Define the population surrogate loss as
\begin{align}
L(\theta) \eqdef \E_{z \sim p^*}[\ell(z,\theta)].
\end{align}
In standard passive learning, we sample $\sD$ randomly from the population
and compute $\theta_\sD$.  As $|\sD| \to \infty$, the parameters generally converge to the minimizer of $L$,
which is in general distinct from the minimizer of $Z$.

\paragraph{Active learning: uncertainty sampling.}
In this work, we consider the streaming setting \citep{settles2010active}
where a learner receives a stream of unlabeled examples (known $x$ drawn from $p^*$ with unknown $y$) and must decide to label each point or not.
We analyze uncertainty sampling in this setting \citep{lewis1994sequential,settles2010active}, which is widely
used for its simplicity and efficacy \citep{yang2016benchmark}.

Let us denote our label budget as $n$, the number of points we label.
Uncertainty sampling (Algorithm \ref{alg:us})
begins with $\nseed < n$ labeled points $\sD$ drawn from the beginning of the stream and
minimizes the regularized loss \refeqn{regerm} to obtain initial parameters.
Then, the algorithm takes a point from the stream and labels it with probability $q(S(x,\theta)/r)) \in [0,1]$ for some acceptance function $q$ and scalar $r$.
One example is $q(s) = \textbf{1}[|s|\leq 1]$, which corresponds to labeling points from the stream if and only if $|S(x,\theta)|\leq r$.
As $r$ gets smaller, we choose points closer to the decision boundary.
\begin{wrapfigure}{r}{.5\linewidth}
  \begin{minipage}[t]{.98\linewidth}

    \begin{algorithm}[H]
      \caption{Uncertainty Sampling}
      \label{alg:us}
      \begin{algorithmic}
        \STATE {\bfseries Input:}
        Regularization parameter $\lambda$,
        label budget $n$,
        seed set size $\nseed$, acceptance function $q$, scale $r$
        \STATE
        \STATE Draw $\nseed$ points from stream, label them, and set them as $\sD$
        \STATE Train $\theta_{\nseed} = \argmin_\theta \sum_{z \in \sD} \ell(z,\theta) + \lambda \|\theta\|_2^2$
        \FOR{$t=(\nseed+1), \dots, n$}
          \LOOP
            \STATE Draw $x$ from unlabeled stream
            \STATE With probability $q(S(x,\theta)/r)$, \textbf{break}
          \ENDLOOP
          \STATE Query $x$ to get label $y$
          \STATE $\sD = \sD \cup \{(x,y)\}$  
          \STATE Train $\theta_t = \argmin_\theta \sum_{z \in \sD} \ell(z,\theta) + \lambda \|\theta\|_2^2$
        \ENDFOR
      \end{algorithmic}
    \end{algorithm}
  \end{minipage}
\end{wrapfigure}
If we decide to label $x$, then we obtain the corresponding label $y$ and add $(x,y)$ to $\sD$.
Finally, we update the model by optimizing \refeqn{regerm}.
The process is continued until we have labeled $n$ points in
total.
\section{Theory}

We present four types of theoretical results. First, in Section \ref{sec:incremental_parameter_update}, we show how the optimal parameters change with the addition of a single point to the convex surrogate (e.g. logistic) loss. Then, in Section \ref{sec:us_updates},
we introduce a smoothed version of the zero-one loss and show that uncertainty sampling is preconditioned stochastic gradient descent on this smoothed zero-one loss. Finally, we show that uncertainty sampling iterates in expectation move in a descent direction in Section \ref{sec:descent_direction}, and that the uncertainty sampling converges to a stationary point of the smoothed zero-one loss (Section \ref{sec:convergence}).

\subsection{Incremental Parameter Updates}
\label{sec:incremental_parameter_update}

First, we analyze how the sample convex surrogate loss minimizer changes with a single additional point,
showing that the change is a preconditioned\footnote{Preconditioned refers to multiplication of a symmetric positive semidefinite matrix by the (stochastic) gradient for (stochastic) gradient descent \citep{li2018preconditioned,klein2011preconditioned}. It is often chosen to approximate the inverse Hessian.} gradient step on the additional point.
Let us assume the loss is convex and thrice differentiable with bounded derivatives:
\begin{assumption}[Convex Loss]
\label{assum:convex_loss}
The loss $\ell(z,\theta)$ is convex in $\theta$.
\end{assumption}
\begin{assumption}[Loss Regularity]
\label{assum:loss_regularity}
The loss $\ell(z,\theta)$ is continuously thrice differentiable in $\theta$, and the first three derivatives are bounded by some constant $\ellbound$ in the Frobenius norm.
\end{assumption}
Consider any iterative algorithm that adds a single point each time and minimizes
the regularized training loss at each iteration $t$:
\begin{align}
\label{eq:loss_with_regularization}
L_t(\theta) \eqdef \sum_{i=1}^t \ell(z^{(i)},\theta) + \lambda \|\theta\|_2^2
\end{align}
to produce $\theta_t$.
Since $L_{t-1}$ and $L_t$ differ by only one point, we expect
$\theta_{t-1}$ and $\theta_t$ to also be close.
We can make this formal using Taylor's theorem.
First, since $\theta_t$ is a minimizer, we have $\nabla L_t(\theta_t) = 0$.
Then, since the loss is continuously twice-differentiable:
\begin{align}
0 = \nabla L_t(\theta_t) = \nabla L_t(\theta_{t-1}) + \left[ \int_0^1 \nabla^2 L_t((1-u) \theta_{t-1} + u \theta_t) du \right]  (\theta_t - \theta_{t-1}) .
\end{align}
Let $P_t$ be the value of the integral.
Since $\ell$ is convex and regularizer is quadratic, $\nabla^2 L_t$ is symmetric positive definite, and thus $P_t$ is symmetric positive definite and thus invertible. Therefore, we can solve for $\theta_t$:
\begin{align}
\theta_t = \theta_{t-1} - P_t^{-1} \nabla L_t(\theta_{t-1}).
\end{align}
Since $\theta_{t-1}$ minimizes $L_{t-1}$, we have $\nabla L_{t-1}(\theta_{t-1})=0$.
Also note that $L_t(\theta) = L_{t-1}(\theta) + \ell(z^{(t)},\theta)$.
Thus,
\begin{align}
\label{eq:precond_sgd}
\theta_t &= \theta_{t-1} - P_t^{-1}  \nabla \ell(z^{(t)},\theta_{t-1}) .
\end{align}
The update above holds for any choice of $z^{(t)}$,
in particular, when $z^{(t)}$ is chosen by either random sampling or uncertainty sampling. 

For random sampling, $z^{(t)} \sim p^*$, so we have
\begin{align}
\label{eq:random_sampling_Ft}
\E[\nabla \ell(z^{(t)},\theta_{t-1})] = \nabla L(\theta_{t-1}),
\end{align}
from which one can interpret the iterates of random sampling as preconditioned SGD on the population surrogate loss $L$.

\subsection{Parameter Updates of Uncertainty Sampling}
\label{sec:us_updates}

Whereas random sampling is preconditioned SGD on the population \emph{surrogate} loss $L$,
we will now show that uncertainty sampling is preconditioned SGD on a smoothed version of the population \emph{zero-one} loss $Z$. 

Recall that $q$ is the acceptance function.
Define $Q$ as a normalized anti-derivative of $q$, which converges to the Heaviside function when the domain is scaled by $r$. First, we need to make an assumption that ensures $q$ has an integral over the real line.

\begin{assumption}[Continuous, bounded, even]
\label{assum:smooth_bounded_q}
The function $q(s)$ 
\begin{itemize}
\item is continuous,
\item has bounded support ($\exists M_q: |s| \geq M_q \implies q(s) = 0$),
\item is even ($q(s)=q(-s)$).
\end{itemize}
\end{assumption}

Now we are ready to define $Z_r$, which is made by replacing the Heaviside step function with $Q$ and scaling the domain by $r$.

\begin{align}
Q_\infty &= \int_{-\infty}^\infty q(u) du \\
Q(s) &= \frac{\int_{-\infty}^s q(u) du}{Q_\infty} \\
Z_r(\theta) &= \E_{(x,y) \sim p^*}[Q(-yS(x,\theta)/r)]
\end{align}

We now show that $Z_r$ converges to $Z$ pointwise:
\begin{proposition}
\label{prop:pointwiseZ}
For all $\theta$,
$\lim_{r \rightarrow 0} Z_r(\theta) = Z(\theta)$.
\end{proposition}
\begin{proof}
This follows from noticing that $\lim_{r \rightarrow 0} Q(s/r) = H(s)$ and applying the Dominated Convergence Theorem.
\end{proof}

Before stating the dynamics of uncertainty sampling, we must first define another quantity, $a_r(\theta)$ which is the probability of accepting a random point $x$:
\begin{align}
a_r(\theta) \eqdef \E[q(S(x,\theta)/r)].
\end{align}

We must further make an assumption that $\psi$ is locally exactly linear around $0$, which is satisfied for the hinge loss and smoothed versions of the hinge loss.
\begin{assumption}[Locally linear $\psi$]
\label{assum:local_linear_psi}
There is some neighborhood of $0$ where $\psi$ is exactly linear:
\begin{align}
\exists m_\psi: |s| \leq m_\psi \implies \psi'(s) = \psi'(0)
\end{align}
\end{assumption}

We also assume the score $S$ is smooth,
and the support of $p^*$ is bounded:
\begin{assumption}[Smooth Score]
\label{assum:smooth_score}
The score $S(x,\theta)$ is smooth, that is, all derivatives with respect to $x$ and $\theta$ exist.
\end{assumption}

\begin{assumption}[Bounded Support]
\label{assum:bounded_support}
The support of $p^*$ is bounded.
\end{assumption}

We are ready to state the relationship between uncertainty sampling iterates (\ref{eq:precond_sgd}), which are governed by $\nabla \ell(z,\theta)$,
and the smoothed zero-one loss $Z_r$:
\begin{theorem}
\label{thm:key_observation}
Under assumptions \ref{assum:smooth_bounded_q}, \ref{assum:local_linear_psi}, \ref{assum:smooth_score}, and \ref{assum:bounded_support},
for any $\theta$,
if $a_r(\theta) \neq 0$ and if $z$ is chosen via uncertainty sampling with parameters $\theta$ and scale $r \leq m_\psi/M_q$,
\begin{align}
  \E[\nabla \ell(z,\theta)] = \eta \nabla Z_r(\theta), \quad\quad \eta \eqdef \frac{-\psi'(0) Q_\infty r}{a_r(\theta)}.
\end{align}
\end{theorem}
\begin{proof}
First, note that
\begin{align}
\nabla Z_r(\theta) = \nabla \int Q(-yS(x,\theta)/r) dp^*(x,y).
\end{align}
Because $Q$ is continuously differentiable, $S$ is smooth, and $p^*$ has bounded support, by the Leibniz Integral Rule, we can exchange the integral and derivative:
\begin{align}
\nabla Z_r(\theta) = \int \frac{q(-yS(x,\theta)/r)}{Q_\infty} (-y/r) \nabla_\theta S(x,\theta) dp^*(x,y).
\end{align}
Because $q$ is even,
\begin{align}
\nabla Z_r(\theta) = - \frac{1}{Q_\infty r} \int q(S(x,\theta)/r) y \nabla_\theta S(x,\theta) dp^*(x,y).
\end{align}
Now we are ready to evaluate $\E[\nabla \ell(z,\theta)]$. From the definition of uncertainty sampling,
\begin{align}
\E[\nabla \ell(z,\theta)] &= \frac{\int q(S(x,\theta)/r) \nabla \ell(z,\theta) dp^*(x,y)}{\int q(S(x,\theta)/r) dp^*(x,y)} \\
&= \frac{1}{a_r(\theta)} \int q(S(x,\theta)/r) \psi'(yS(x,\theta)) y \nabla_\theta S(x,\theta) dp^*(x,y)
\end{align}
Notice that $q(S(x,\theta)/r)=0$ for $|S(x,\theta)|\geq r M_q$ and that $\psi'(s) = \psi'(0)$ for $|s| \leq m_\psi$. Thus, for $r \leq m_\psi / M_q$,
\begin{align}
\E[\nabla \ell(z,\theta)] &= \frac{1}{a_r(\theta)} \int q(S(x,\theta)/r) \psi'(0) y \nabla_\theta S(x,\theta) dp^*(x,y) \\
&= \frac{-\psi'(0) Q_\infty r}{a_r(\theta)} \nabla Z_r(\theta).
\end{align}
\end{proof}

Thus, if $z^{(t)}$ is drawn using uncertainty sampling, $\E[\nabla \ell(z^{(t)},\theta_{t-1})]$ is in the direction of $\nabla Z_r(\theta_{t-1})$, since the quantity in front of $\nabla Z_r$ is a scalar that is positive for all common losses.\footnote{$Q_\infty$, $r$, and $a_r(\theta)$ are all positive, and $\psi'(0)$ is negative for all reasonable losses.}
Similar to how we showed random sampling is preconditioned SGD on the population surrogate loss $L$ (\ref{eq:random_sampling_Ft}),
uncertainty sampling is preconditioned SGD on the smoothed population zero-one loss $Z_r$.

The only assumption that is unorthodox is Assumption \ref{assum:local_linear_psi},
which holds for a smoothed hinge loss, but not the logistic loss.
If we remove this assumption, we would incur only a small additive vector term of $O(r^2 / a_r(\theta))$ which goes to $0$ quickly as $r \rightarrow 0$.

While $Z_r(\theta) \to Z(\theta)$ pointwise as $r \to 0$ (Proposition~\ref{prop:pointwiseZ}),
showing that $\nabla Z_r (\theta) \rightarrow \nabla Z(\theta)$ requires additional assumptions (see Section \ref{sec:convergence_of_gradient} in the Appendix for more details).

\subsection{Descent Direction}
\label{sec:descent_direction}

So far, we have shown that uncertainty sampling is preconditioned SGD on the smoothed
population zero-one loss $Z_r$ by analyzing $\E[\nabla \ell(z, \theta)]$.
To show that these updates are descent directions on $Z_r$,
we need to also consider the preconditioner $P_t^{-1}$
appearing in (\ref{eq:precond_sgd}).
Due to quadratic regularization (\ref{eq:loss_with_regularization}),
the preconditioner is positive definite.
However, we need to be careful since the preconditioner depends on the resulting iterate $\theta_t$.
Because of this snag,
we need to ensure that $\theta_t$ doesn't change the preconditioner $P_t^{-1}$ too much,
which we can accomplish by requiring $\|\nabla Z(\theta_{t-1})\|\geq \epsilon$ and large enough regularization.

\begin{theorem}[Uncertainty Sampling Descent Direction]
\label{thm:us_descent}
Assume that Assumptions \ref{assum:convex_loss}, \ref{assum:loss_regularity}, \ref{assum:smooth_bounded_q}, \ref{assum:local_linear_psi}, \ref{assum:smooth_score}, and \ref{assum:bounded_support} hold, and $\psi'(0) < 0$. For any $\epsilon>0$ and $n$, for any
\begin{align}
\lambda \geq \max \left( \sqrt{\frac{4 M_\ell^3 n}{\epsilon (-\psi'(0))Q_\infty r}} , \sqrt[3]{\frac{M_\ell^4 n^2}{\epsilon (-\psi'(0)) Q_\infty r}} \right),
\end{align}
for all iterates of uncertainty sampling $\{\theta_t\}$ such that $\|\nabla Z(\theta_{t-1}) \| \geq \epsilon$, we have
\begin{align}
\nabla Z_r(\theta_{t-1}) \cdot \mathbb{E}[\theta_t - \theta_{t-1}|\theta_{t-1}] < 0 .
\end{align}
\end{theorem}

It might seem that $\lambda = \Omega(n^{2/3})$ above is rather large.
However, note that we don't optimize the regularized loss until after $\nseed$ random points,
so if $\nseed \geq \lambda$, then the regularization is always less than the number of data points that contribute to the loss, and the regularization will not dominate. Thus, this constraint on $\lambda$ can be intuitively thought of as a constraint on $\nseed = \Omega(n^{2/3})$.

\subsection{Convergence}
\label{sec:convergence}

Having shown that uncertainty sampling iterates move in descent directions of $Z_r$ in expectation,
we now turn to showing that they also converge to a stationary point of $Z_r$. 
To prove convergence, we will need to
stay (with high probability) in regions where the assumption of Theorem \ref{thm:key_observation} holds ($a_r(\theta) \neq 0$)
and also ensure that the parameters stay bounded (with high probability).

As is standard in stochastic gradient convergence analyses,
instead of showing convergence of the final parameter iterate $\theta_n$,
we show the convergence of the parameters from a random iteration.
For a budget $n$, let $\tilde{t} \in [\nseed,n)$ with $\BP(\tilde{t} = t) \propto 1/t$.
Then, define $\tilde{\theta}_n \eqdef \theta_{\tilde{t}}$ be the \emph{randomized parameters}.
We will show that $\tilde \theta_n$ converges to a stationary point of $Z_r$ as $n \rightarrow \infty$.

Define $\delta_{C,r,n}$ as the failure probability that any parameter iterate $\theta_t$ is too large or has zero acceptance probability:
\begin{align}
\delta_{C,r,n} \eqdef \mathbb P[\exists t \in [\nseed,n]: \|\theta_t\| > C \vee a_r(\theta_t)=0].
\end{align}
We will assume that this failure probability converges to $0$ as $n \rightarrow \infty$. As $n$ grows, the seed set size grows, the regularization grows, and the effective step size shrinks. Intuitively, this means we might expect that the parameter iterates become more stable. Unless the parameters diverge, we can choose $C$ large enough to contain the parameter iterates. Furthermore, the acceptance probability should be non-zero if there are not large regions of the space with zero probability density. Assuming the probability of these failure events goes to zero,
the randomized parameters converge to a stationary point:
\begin{theorem}[Convergence to Stationary Points]
\label{thm:convergence}
Assume that Assumptions \ref{assum:convex_loss}, \ref{assum:loss_regularity}, \ref{assum:smooth_bounded_q}, \ref{assum:local_linear_psi}, \ref{assum:smooth_score}, \ref{assum:bounded_support} hold
and that $\psi'(0) < 0$.
Assume $\lambda = \Omega(n^{2/3})$ and $\nseed = o(n)$,
and there exists some $C>0$ such that $\delta_{C,r,n} \rightarrow 0$ as $n \rightarrow \infty$.
Then for the randomized parameters produced by uncertainty sampling with fixed scale $r \leq m_\psi/M_q$,
  we have that as $n \rightarrow \infty$,
\begin{align}
\nabla Z_r(\tilde{\theta}_n) \stackrel{P}{\rightarrow} 0.
\end{align}
\end{theorem}

These results shed light on the mysterious dynamics of uncertainty sampling which motivated this paper.
In particular, uncertainty sampling can achieve lower zero-one loss than random sampling
because it is implicitly descending on the smoothed zero-one loss $Z_r$.
Furthermore, since $Z_r$ is non-convex,
uncertainty sampling can converge to different values depending on the initialization.

It is important to note that the actual uncertainty sampling algorithm is unchanged---it is still
performing gradient updates on the convex surrogate loss.
But because its sampling distribution is skewed towards the decision boundary,
we can interpret its updates as being on the smoothed zero-one loss
with respect to the original data-generating distribution.

\section{Experiments}

We run uncertainty sampling on a simple synthetic dataset to illustrate the dynamics (\refsec{synthetic})
as well as 22 real datasets (\refsec{real}). 
In both cases, we show how uncertainty sampling converges to different parameters depending on initialization,
and how it can achieve lower asymptotic zero-one loss compared to minimizing the surrogate loss on all the data.
Note that most active learning experiments are interested in measuring the rate
of convergence (data efficiency), whereas this paper focuses exclusively on asymptotic values and the
variation that we obtain from different seed sets.
We evaluate only on the zero-one loss, but all algorithms perform optimization on the logistic loss.

\subsection{Synthetic Data}
\label{sec:synthetic}

\begin{figure}
    \centering
    \begin{minipage}{.5\textwidth}
        \centering
        \includegraphics[width=0.7\linewidth]{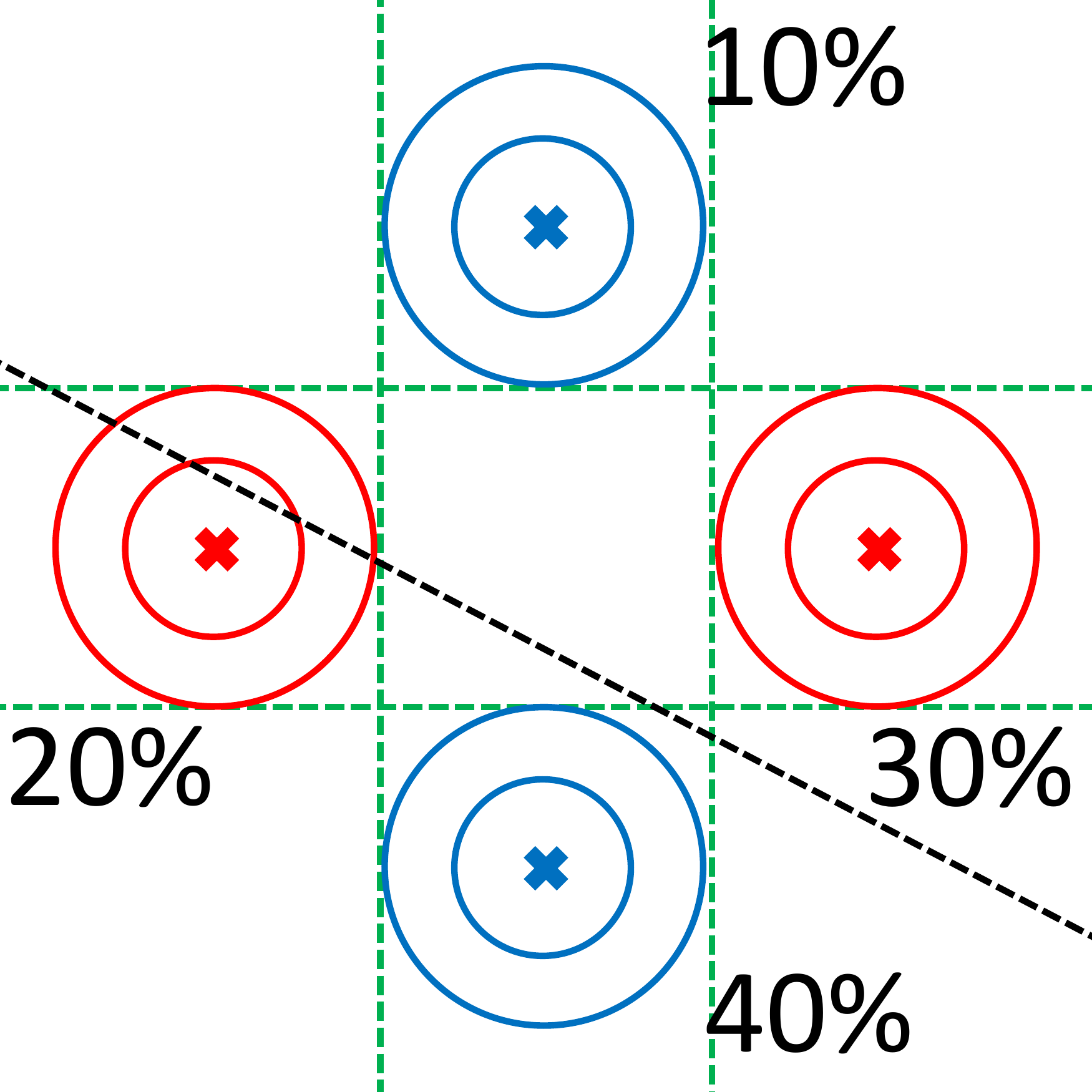}
    \end{minipage}%
    \hfill
    \begin{minipage}{0.5\textwidth}
        \centering
        \includegraphics[width=1.0\linewidth]{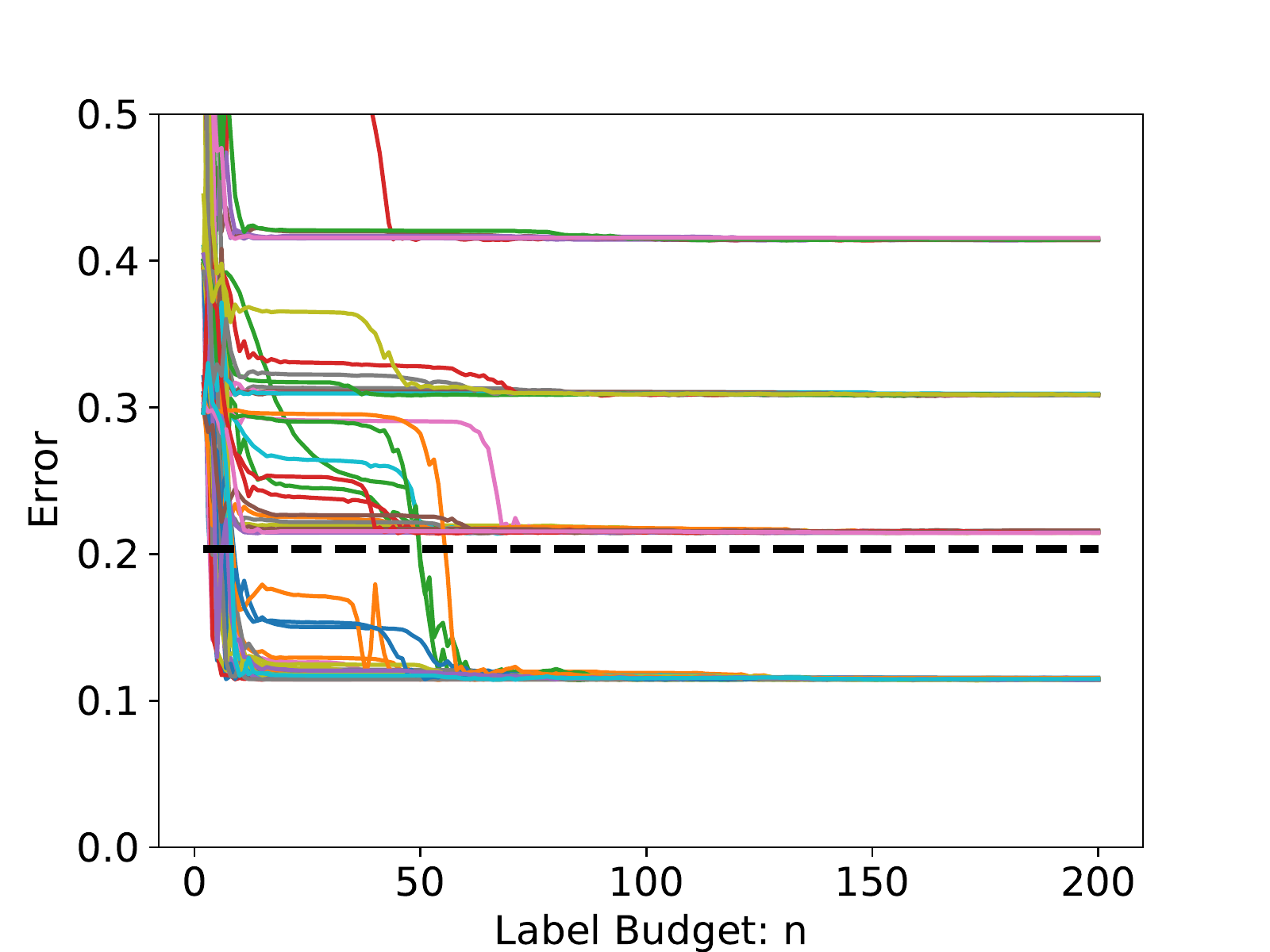}
        
        \label{fig:synthetic1}
    \end{minipage}
    \caption{Synthetic dataset based on a mixture of four Gaussians (left)
    and the associated learning curves for runs of uncertainty sampling with different initial seed sets (right).
    Depending on the seed set, uncertainty sampling can produce either better or worse parameters
    than random sampling.
     \label{fig:gaussian}}
\end{figure}

Figure~\ref{fig:gaussian} (left) shows a mixture of
Gaussian distributions in two dimensions. All the Gaussians are isotropic, and
the size of the circle indicates the variance (one standard deviation for the
inner circle, and two standard deviations for the outer circle).
The points drawn from the two red Gaussian distributions are labeled $y=1$
and the points drawn from the two blue ones are labeled $y=-1$. The
percentages refer to the mixture proportions of the clusters.
We see that there are four local minima of the population zero-one loss,
indicated by the green dashed lines. Each minimum misclassifies one of the Gaussian clusters,
yielding error rates of about 10\%, 20\%, 30\%, and 40\%.
The black dotted line corresponds to the parameters that minimize the logistic loss,
which yields an error of about 20\%.

Figure~\ref{fig:gaussian} (right) shows learning curves for different seed sets,
which consist of two points, one from each class. We see
that the uncertainty sampling learning curves converge to four different asymptotic losses,
corresponding to the four local minima of the zero-one loss mentioned earlier.
The thick black dashed line is the zero-one loss for random sampling.
We see that uncertainty sampling can actually achieve lower loss than random sampling,
since the global optimum of the logistic loss does not correspond to the global minimum of the zero-one loss.

\subsection{Real-World Datasets}
\label{sec:real}

We collected 22 datasets from OpenML (retrieved August, 2017) that had a large
number of data points and where logistic regression outperformed the baseline
classifier that always predicts the majority label. We further subsampled each dataset to
have 10,000 points, which was divided into 7000 training points and 3000 test
points.
We created a stream of points by randomly selecting points one-by-one with replacement from the dataset.
We ran uncertainty sampling on each dataset with random seed sets of sizes that are powers of
two from 2 to 4096 and then 7000.
We stopped when uncertainty sampling either could not select a point ($a_r(\theta)=0$) or when a point had been repeatedly selected more than $10$ times.
For each dataset and seed set size, we ran uncertainty sampling 10 times, for a total of
$22 \cdot 13 \cdot 10 = 2860$ runs.

\begin{figure}
    \centering
    \begin{minipage}{.48\textwidth}
        \centering
        \includegraphics[width=1.0\linewidth]{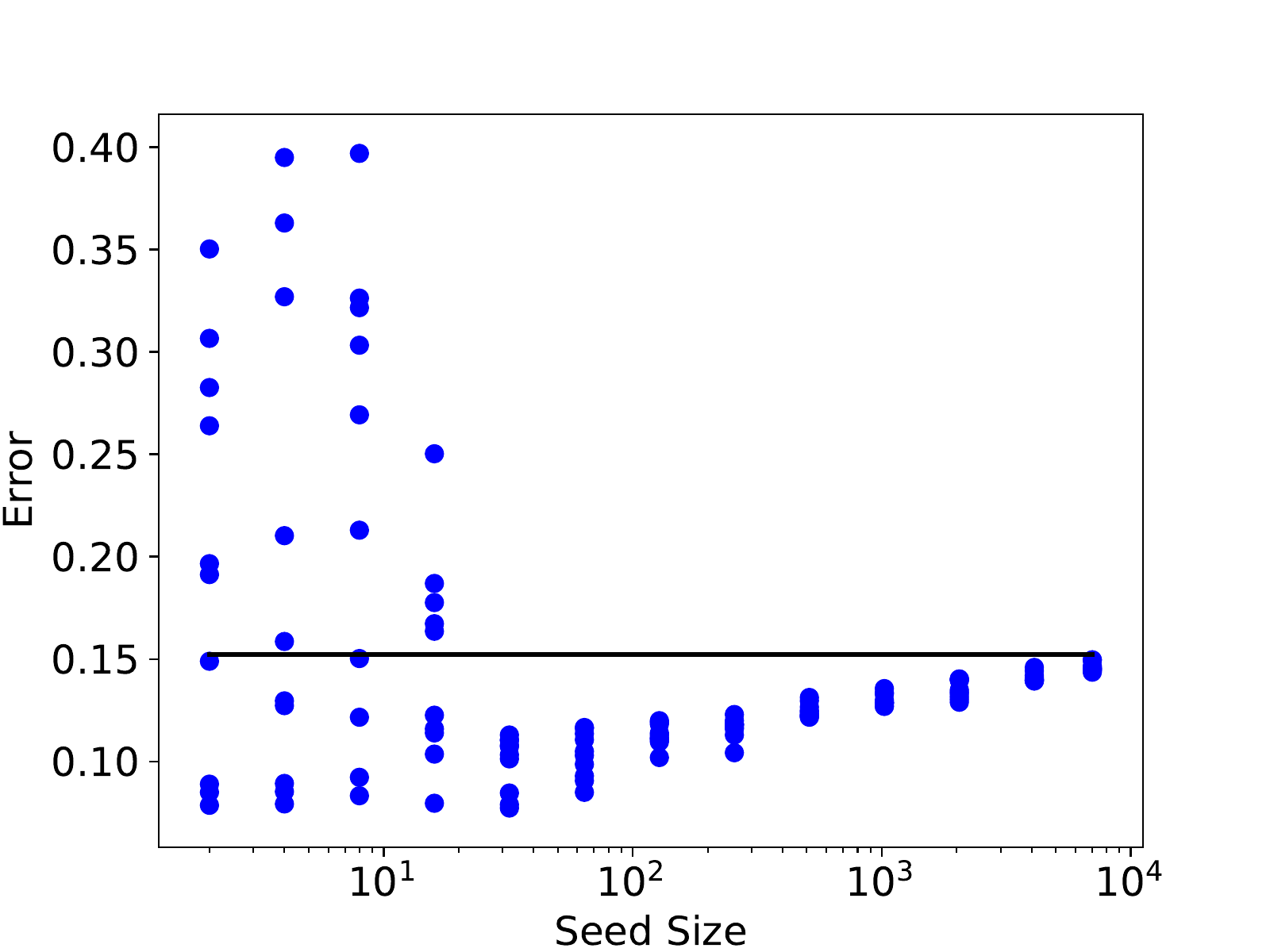}
    \end{minipage}%
    \hfill
    \begin{minipage}{0.48\textwidth}
        \centering
        \includegraphics[width=1.0\linewidth]{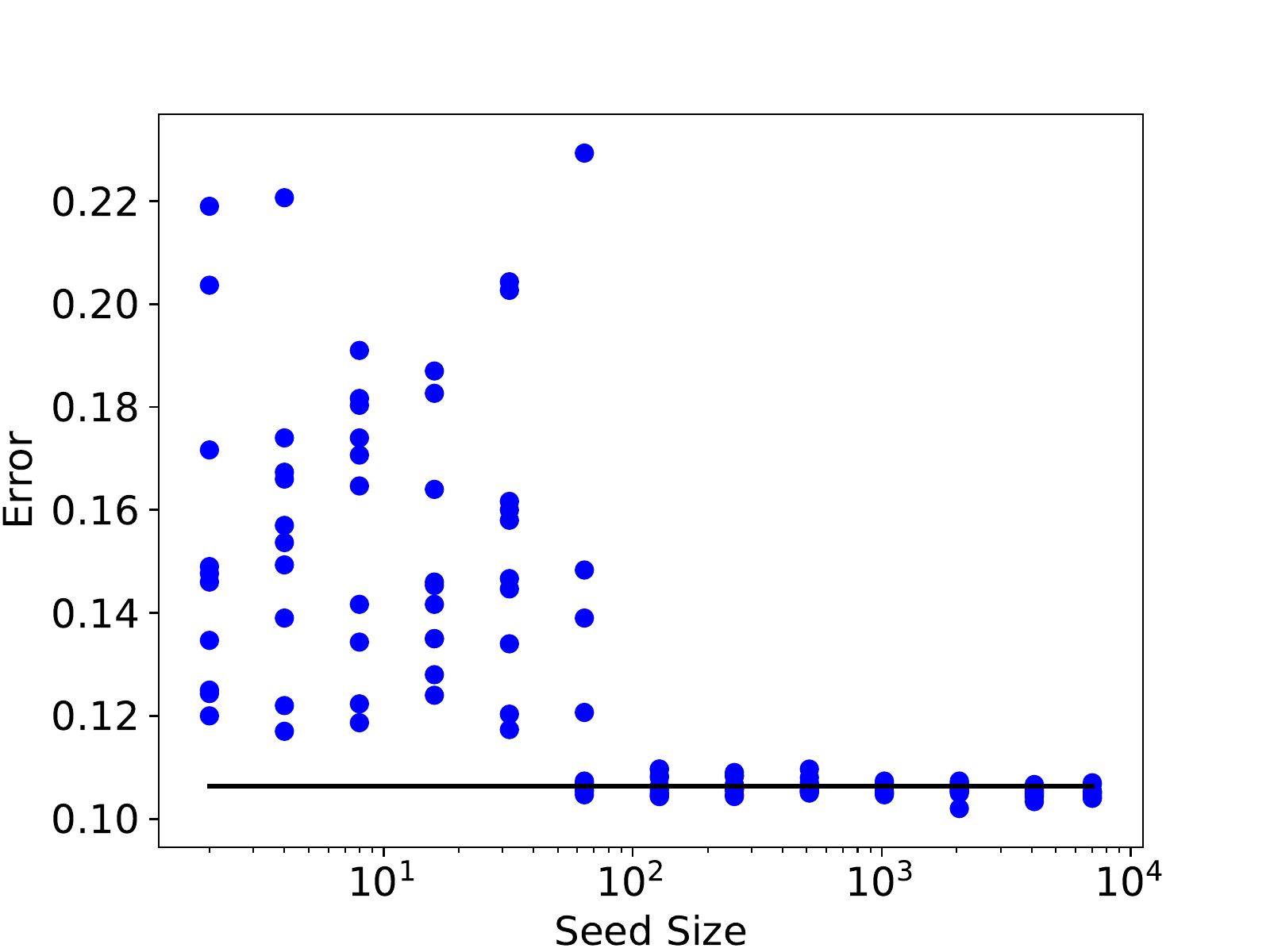}
    \end{minipage}
    \caption{A scatter plot of the asymptotic zero-one loss for uncertainty sampling for two particular datasets for 13 seed sizes. The black line is the zero-one loss on the full dataset.
    }
    \label{fig:scatter}
\end{figure}

In Figure \ref{fig:scatter}, we see scatter plots of the asymptotic zero-one loss
of 130 points: 13 seed set sizes, each with 10 runs. The dataset on the left was chosen to
exhibit the wide range of convergence values of uncertainty sampling, some with lower zero-one loss than with the full dataset.
In both plots, we see that the variance of the zero-one loss of uncertainty sampling decreases as the seed set grows.
This is expected from theory since the initialization has less variance for larger seed set sizes (as the seed set size goes to infinity, the parameters converge).
For most of the datasets, the behavior was more similar to
the plot on the right, where uncertainty sampling has a higher mean zero-one loss than random sampling for most seed sizes.

\begin{figure}
    \centering
    \begin{minipage}{.48\textwidth}
        \centering
        \includegraphics[width=1.0\linewidth]{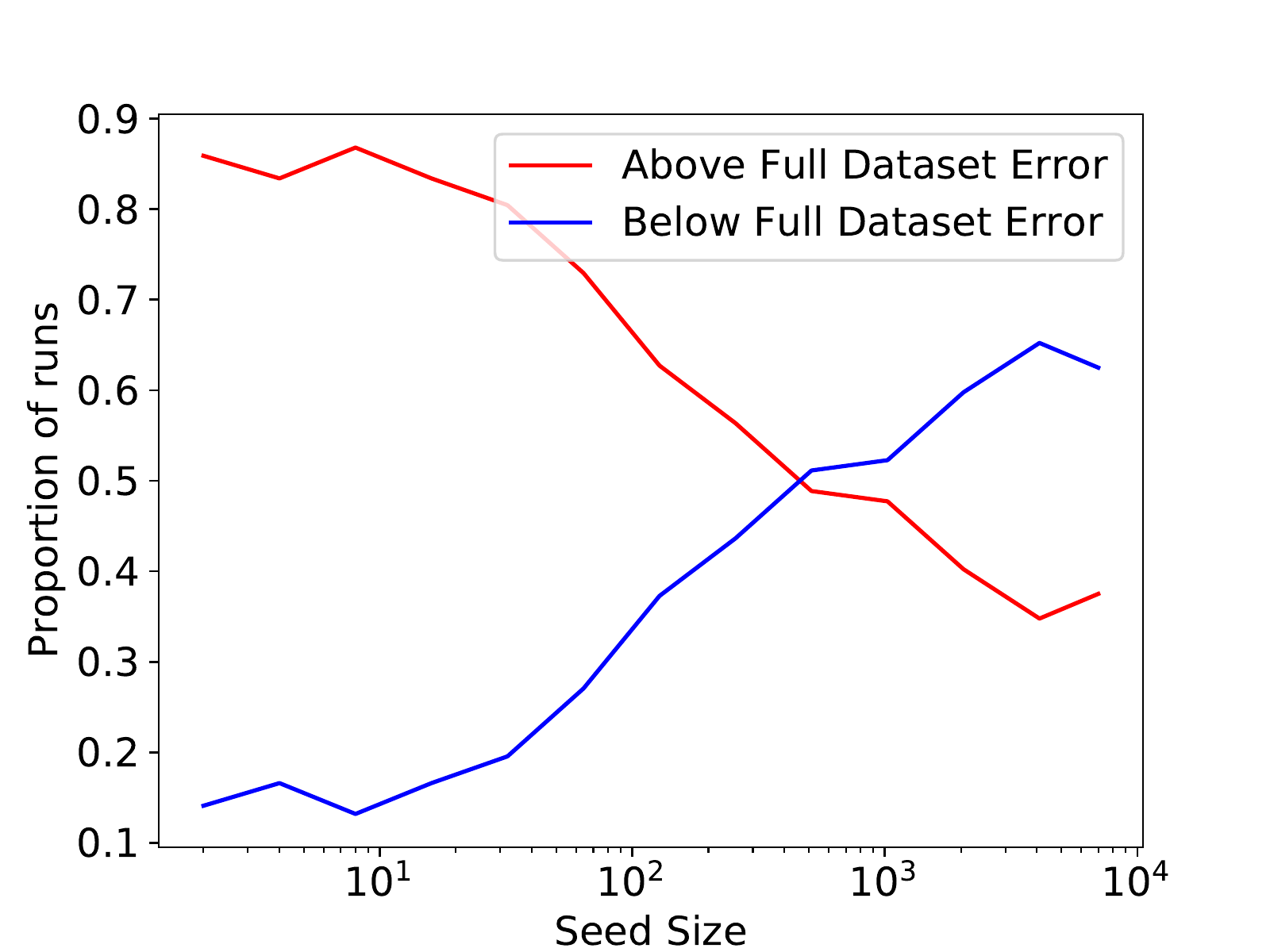}
        \caption{A plot showing the distribution of runs over the datasets
        (with 10 runs per dataset) of when uncertainty sampling converges to a lower zero-one loss than using the entire dataset.
        }
        \label{fig:proportion_plot}
    \end{minipage}%
     \hfill
    \begin{minipage}{0.48\textwidth}
        \centering
        \includegraphics[width=1.0\linewidth]{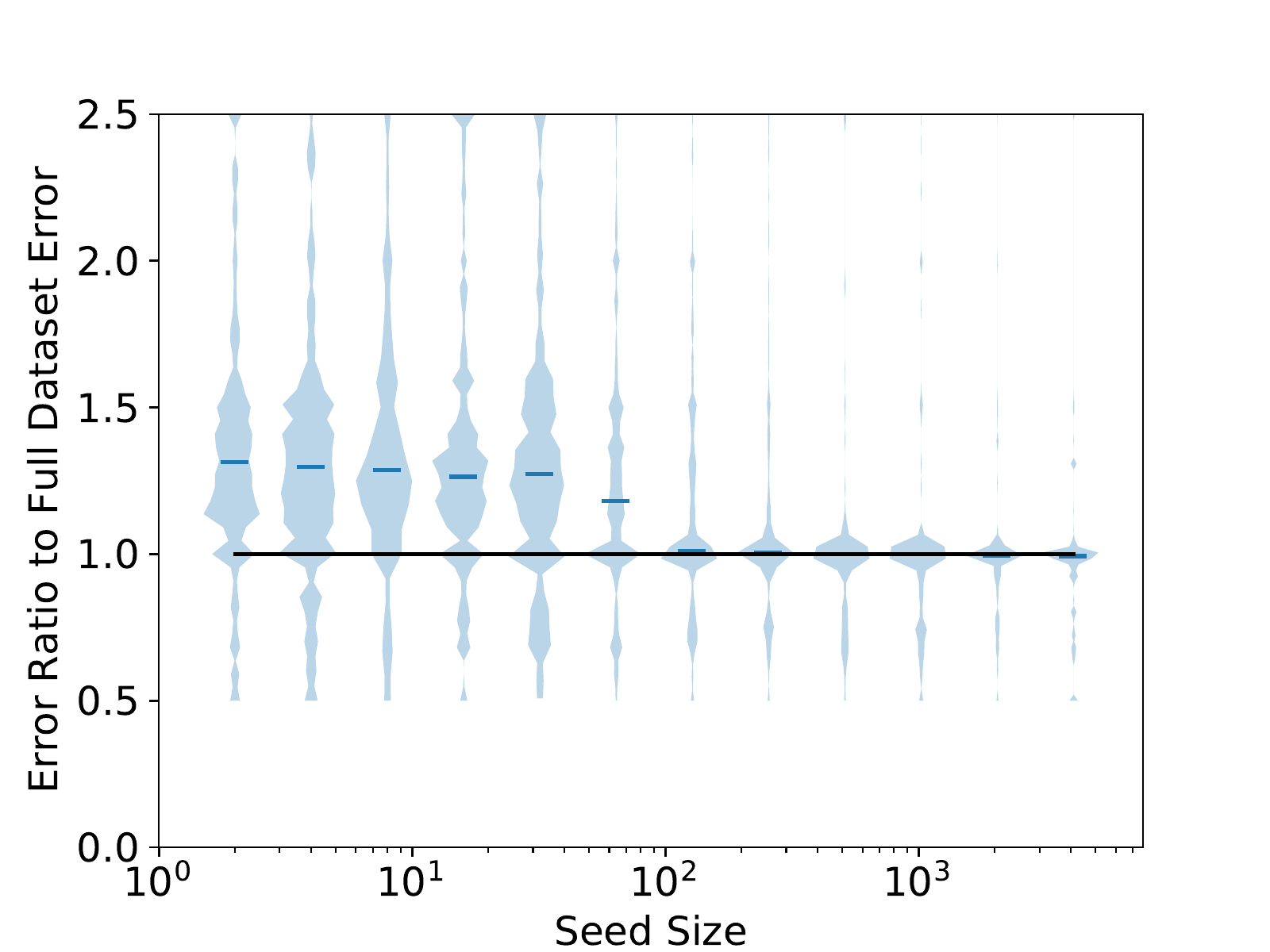}
        \caption{A violin plot capturing the relative asymptotic zero-one loss
        compared to the zero-one loss on the full dataset. The plot shows the
        density of points with kernel density estimation. The red lines are the
        median losses. Each ``violin'' captures 220 points (10 runs over 22 datasets).}
        \label{fig:violin_plot}
    \end{minipage}
\end{figure}

To gain a more quantitative understanding of all the datasets, we summarized the
asymptotic zero-one loss of uncertainty sampling for various random seed set sizes.
In Figure \ref{fig:proportion_plot}, we show the proportions of the runs over
the datasets where uncertainty sampling converges to a lower zero-one loss
than using the entire dataset. In Figure \ref{fig:violin_plot}, we show a ``violin plot'' for the
distribution of the ratio between the asymptotic zero-one loss of uncertainty
sampling and the zero-one loss using the full dataset. We note that the mean and variance
of uncertainty sampling significantly drops as the size of the seed set
grows larger.
The initial parameters are poor if the seed set is small, and it is well-known that
poor initializations for optimizing non-convex functions locally can yield poor
results, as seen here.

\section{Related Work and Discussion}

The phenomenon that uncertainty sampling can achieve lower error with a subset
of the data rather than using the entire dataset has been observed multiple
times in the literature. In fact, the original uncertainty sampling paper
\citep{lewis1994sequential} notes that ``For 6 of 10 categories, the mean
[F-score] for a classifier trained on a uncertainty sample of 999 examples
actually exceeds that from training on the full training set of 319,463''.
\citet{schohn2000less} defines a heuristic that selects the point closest to
the decision boundary of an SVM, which is similar to uncertainty sampling in
our formulation. In the abstract, the authors note, ``We observe... that a SVM
trained on a well-chosen subset of the available corpus frequency performs
better than one trained on \emph{all} available data''. More recently,
\citet{chang2017active} developed an ``active bias'' technique that emphasizes
the uncertain points and found that it increases the performance compared to using a
fully-labeled dataset. 

There is also work showing the bias of active learning can harm final performance.
\citet{schutze2006performance} notes the ``missed cluster effect'', where active learning can ignore clusters in the data and never query points from there; this is seen in our synthetic experiment. \citet{dasgupta2008hierarchical} has a section on the bias of uncertainty sampling and provides another example where uncertainty sampling fails due to sampling bias, which we can explain as due to local minima of the zero-one loss.
\citet{bach2007active} and \citet{beygelzimer2009importance} note this bias issue and propose different importance sampling schemes to re-weight points and correct for the bias.  

In this work, we showed that uncertainty sampling updates are preconditioned SGD steps on the population zero-one loss and move in descent directions for parameters that are not approximate stationary points.
Note that this does not give any global optimality guarantees.
In fact, for linear classifiers, it is NP-hard to optimize the training
zero-one loss below $\frac{1}{2}-\epsilon$ (for any $\epsilon>0$) even when
there is a linear classifier that achieves just $\epsilon$ training zero-one
loss \citep{feldman2012agnostic}. 

One of the key questions in light of this work is when optimizing convex surrogate
losses yield good zero-one losses. If the loss function corresponds to the negative log-likelihood of a
\emph{well-specified} model, then the zero-one loss $Z$ will have a local minimum at the
parameters that optimize the log-likelihood.
If the loss function is ``classification-calibrated'' (which holds for most common surrogate losses),
\citet{bartlett2006convexity} shows that if the convex surrogate loss of the estimated parameters
converges to the optimal convex surrogate loss, then the zero-one loss of the
estimated parameters converges to the global minimum of the zero-one loss (Bayes error).
This holds only for universal classifiers \citep{micchelli2006universal},
and in practice, these assumptions are unrealistic. For instance, several
papers show how outliers and noise can cause linear classifiers learned on convex surrogate losses to suffer high
zero-one loss \citep{nguyen2013algorithms,wu2007robust,long2010random}.

Other works connect active learning with optimization in rather different ways. \citet{ramdas2013algorithmic} uses active learning as a subroutine to improve stochastic convex optimization.
\citet{guillory2009active} shows how performing online active learning updates corresponds to online optimization updates of non-convex functions, more specifically, truncated convex losses. In this work, we analyze active learning with offline optimization and show the connection between uncertainty sampling and one particularly important non-convex loss, the zero-one loss.

In summary, our work is the first to show a connection between the zero-one loss and the commonly-used
uncertainty sampling. This provides an explanation and understanding of
the various empirical phenomena observed in the active learning literature.
Uncertainty sampling simultaneously offers the hope of converging to lower error and the
danger of converging to local minima (an issue that can possibly be avoided with larger seed sizes).
We hope this connection can lead to improved active learning and optimization algorithms.

\paragraph{Reproducibility.}
The code, data, and experiments for this paper
are available on the CodaLab platform at
\href{https://worksheets.codalab.org/worksheets/0xf8dfe5bcc1dc408fb54b3cc15a5abce8/}{https://worksheets.codalab.org/worksheets/0xf8dfe5bcc1dc408fb54b3cc15a5abce8/}.

\paragraph{Acknowledgments.}
\label{sec:acknowledgments}
This research was supported by an NSF Graduate Fellowship to the first author. 

\bibliography{bibliography}
\bibliographystyle{icml2018}

\newpage
\section{Appendix}

The appendix has two main sections. In Section \ref{sec:descent_direction_and_convergence}, we prove the results about the descent direction and SGD convergence of uncertainty sampling. In Section \ref{sec:convergence_of_gradient}, we show that under some conditions, $\nabla Z_r \rightarrow \nabla Z$.

\subsection{Descent Direction and Convergence}
\label{sec:descent_direction_and_convergence}

We prove two lemmas about the parameter updates. 
First, we show that a single step of the parameter iterates don't change much due to regularization (Lemma \ref{lem:bounded_step}). Second, we show that $\theta_t - \theta_{t-1}$ is approximately equal to $- [\nabla^2 L_{t-1}(\theta_{t-1}]^{-1} \nabla \ell(z^{(t)},\theta_{t-1})$ (Lemma \ref{lem:approx_step}) and bound the error in approximation. This is important because it takes the dependency on iterate $t$ only through the gradient of the loss at $z^{(t)}$, the point selected at iterate $t$. With these two lemmas and Theorem \ref{thm:key_observation}, the descent direction (Theorem \ref{thm:us_descent}) is straightforward and the SGD convergence (Theorem \ref{thm:convergence}) follows a standard SGD convergence argument.

\subsubsection{Parameter Update Lemmas}

\begin{lemma}
\label{lem:bounded_step}
\begin{align}
\|\theta_t - \theta_{t-1}\| \leq \frac{M_\ell}{\lambda}
\end{align}
\end{lemma}
\begin{proof}
As in the main text, we have

\begin{align}
L_t(\theta) = \sum_{i=1}^t \ell(z^{(i)},\theta) + \lambda \|\theta\|_2^2.
\end{align}

Thus, $L_t(\theta) = L_{t-1}(\theta) + \ell(z^{(t)},\theta)$ and further $\nabla L_t(\theta_t) = 0$. Together, this implies that $\nabla L_{t}(\theta_{t-1}) = \nabla \ell(z^{(t)},\theta_{t-1})$.

Using the Taylor expansion,

\begin{align}
0 = \nabla L_t(\theta_t) &= \nabla \ell(z^{(t)},\theta_{t-1}) + P_t (\theta_t - \theta_{t-1}),
\end{align}

where
\begin{align}
P_t &= \int_0^1 \nabla^2 L_t((1-u) \theta_{t-1} + u \theta_t) du.
\end{align}

Since the loss is convex with quadratic regularization,

\begin{align}
\lambdamin(P_t) &\geq \int_0^1 \lambdamin(\nabla^2 L_t((1-u) \theta_{t-1} + u \theta_t) du \\
&\geq \int_0^1 \lambdamin(\lambda I) du \\
&\geq \lambda \\
\|P_t^{-1}\| &\leq \frac{1}{\lambda}
\end{align}

Therefore,

\begin{align}
\theta_t - \theta_{t-1} &= -[P_t]^{-1} \nabla \ell(z^{(t)},\theta_{t-1}) \\
\|\theta_t - \theta_{t-1}\| &\leq \frac{M_{\ell}}{\lambda }
\end{align}
\end{proof}


\begin{lemma}
\label{lem:approx_step}
\begin{align}
\|\theta_t - \theta_{t-1} + [\nabla^2 L_{t-1}(\theta_{t-1}]^{-1} \nabla \ell(z^{(t)},\theta_{t-1})\| \leq \frac{M_\ell^2}{\lambda^2} + \frac{\ellbound^3 n}{2 \lambda^3}
\end{align}
\end{lemma}
\begin{proof}
From a Taylor expansion,

\begin{align}
0 = \nabla L_t(\theta_t) &= \nabla \ell(z^{(t)},\theta_{t-1}) + \nabla^2 L_t(\theta_{t-1}) (\theta_t - \theta_{t-1}) + Q,
\end{align}

where 

\begin{align}
Q_i &= (\theta_t - \theta_{t-1})^T \left[ \int_0^1 (1-u) [\nabla^3 L_t(u \theta_t + (1-u) \theta_{t-1} )]_i du \right] (\theta_t - \theta_{t-1}).
\end{align}

We want to solve for $\theta_t - \theta_{t-1}$, but in order to do this, we need to bound $Q$.

\begin{align}
\|Q\| &\leq \|\theta_t - \theta_{t-1}\| \left[ \int_0^1 (1-u) \|\nabla^3 L_t(u \theta_t + (1-u) \theta_{t-1} )\|_F du \right]  \|\theta_t - \theta_{t-1}\|
\end{align}

Using Lemma \ref{lem:bounded_step}
\begin{align}
&\leq \frac{M_\ell}{\lambda} \int_0^1 (1-u) du M_\ell t \frac{M_\ell}{\lambda} \\
&\leq \frac{\ellbound}{\lambda} \frac{n M_\ell}{2} \frac{\ellbound}{\lambda} \\
 &\leq \frac{\ellbound^3 n}{2 \lambda^2}
\end{align}

Solving for $\theta_t - \theta_{t-1}$ in the Taylor expansion,

\begin{align}
\theta_t - \theta_{t-1} &= - [\nabla^2 L_t(\theta_{t-1})]^{-1} (\nabla \ell(z^{(t)},\theta_{t-1}) + Q) \\
\theta_t - \theta_{t-1} + [\nabla^2 L_t(\theta_{t-1})]^{-1} \nabla \ell(z^{(t)},\theta_{t-1}) &= -[\nabla^2 L_t(\theta_{t-1})]^{-1} Q \\
\|\theta_t - \theta_{t-1} + [\nabla^2 L_t(\theta_{t-1})]^{-1} \nabla \ell(z^{(t)},\theta_{t-1})\| &\leq \|[\nabla^2 L_t(\theta_{t-1})]^{-1}\| \|Q\| \\
&\leq \frac{1}{\lambda} \frac{\ellbound^3 n}{2 \lambda^2} \\
&\leq \frac{\ellbound^3 n}{2 \lambda^3} \\
\end{align}

Looking at the theorem statement, we are almost done. The only difference between the theorem statement and the equation above is that the theorem statement has $\nabla^2 L_{t-1}$ while the equation above has $\nabla^2 L_t$. We can use the triangle inequality and bound the difference.

\begin{align}
& \|\theta_t - \theta_{t-1} + [\nabla^2 L_{t-1}(\theta_{t-1})]^{-1} \nabla \ell(z^{(t)},\theta_{t-1})\| \\
&\leq \|[\nabla^2 L_{t-1}(\theta_{t-1})]^{-1} - [\nabla^2 L_t(\theta_{t-1})]^{-1}\| \|\nabla \ell(z^{(t)},\theta_{t-1})\| + \frac{\ellbound^3 n}{2 \lambda^3} \\
&\leq \|[\nabla^2 L_{t-1}(\theta_{t-1})]^{-1} [\nabla^2 L_t(\theta_{t-1}) - \nabla^2 L_{t-1}(\theta_{t-1})][\nabla^2 L_t(\theta_{t-1})]^{-1}\| M_\ell + \frac{\ellbound^3 n}{2 \lambda^3} \\
&\leq \frac{1}{\lambda} M_\ell \frac{1}{\lambda} M_\ell + \frac{\ellbound^3 n}{2 \lambda^3} \\
&\leq \frac{M_\ell^2}{\lambda^2} + \frac{\ellbound^3 n}{2 \lambda^3}
\end{align}

\end{proof}


\subsubsection{Descent Direction}

\begin{thm}[\ref{thm:us_descent}]
Assume that Assumptions \ref{assum:convex_loss}, \ref{assum:loss_regularity}, \ref{assum:smooth_bounded_q}, \ref{assum:local_linear_psi}, \ref{assum:smooth_score}, and \ref{assum:bounded_support} are met. Further, assume $\psi'(0) < 0$. For any $\epsilon>0$ and $n$, for any sufficiently large 
\begin{align}
\lambda \geq \max \left( \sqrt{\frac{4 M_\ell^3 n}{\epsilon (-\psi'(0))Q_\infty r}} , \sqrt[3]{\frac{M_\ell^4 n^2}{\epsilon (-\psi'(0)) Q_\infty r}} \right),
\end{align}
for all iterates of uncertainty sampling $\{\theta_t\}$ such that $\|\nabla Z(\theta_{t-1}) \| \geq \epsilon$, then,
\begin{align}
\nabla Z_r(\theta_{t-1}) \cdot \mathbb{E}[\theta_t - \theta_{t-1}|\theta_{t-1}] < 0 .
\end{align}
\end{thm}
\begin{proof}

The first thing to note is that if $\|\nabla Z_r(\theta_{t-1})\| > 0$, then $a_r(\theta_{t-1})>0$.

\begin{align}
\|\nabla Z_r(\theta_{t-1})\| &= \| \frac{1}{Q_\infty r} \int q(S(x,\theta)/r) y \nabla_\theta S(x,\theta) dp^*(x,y) \| \\
&\leq \frac{1}{Q_\infty r} \int q(S(x,\theta)/r) \| \nabla_\theta S(x,\theta) \| dp^*(x,y)
\end{align}

Because $p^*$ has bounded support and since $S$ is smooth, there exists a constant $C_\theta$ such that $\|\nabla S(x,\theta)\| \leq C_\theta$.

\begin{align}
\|\nabla Z_r(\theta_{t-1})\| &\leq \frac{C_{\theta_{t-1}}}{Q_\infty r} a_r(\theta) .
\end{align}

And thus, if $\|\nabla Z_r(\theta_{t-1})\| \geq \epsilon > 0$, then $a_r(\theta)>0$.

This will allow us to use Theorem \ref{thm:key_observation} later in the proof.

Using Lemma \ref{lem:approx_step},

\begin{align}
& \nabla Z_r(\theta_{t-1}) \cdot (\theta_t - \theta_{t-1}) \\
&\leq - \nabla Z_r(\theta_{t-1})^T [\nabla^2 L_{t-1}(\theta_{t-1})]^{-1} \nabla \ell(z^{(t)},\theta_{t-1}) + \|\nabla Z_r(\theta_{t-1})\| \left( \frac{M_\ell^2}{\lambda^2} + \frac{\ellbound^3 n}{2 \lambda^3}\right).
\end{align}

Note that the only part that is dependent on the $t^{th}$ iteration is the $\nabla \ell$ term, which we can evaluate on the right by Theorem \ref{thm:key_observation}. Thus,

\begin{align}
& \nabla Z_r(\theta_{t-1}) \cdot \E[\theta_t - \theta_{t-1}| \theta_{t-1}] \\
&\leq - \nabla Z_r(\theta_{t-1})^T [\nabla^2 L_{t-1}(\theta_{t-1})]^{-1} \frac{-\psi'(0) Q_\infty r}{a_r(\theta_{t-1})} \nabla Z_r(\theta_{t-1}) + \|\nabla Z_r(\theta_{t-1})\| \left( \frac{M_\ell^2}{\lambda^2} + \frac{\ellbound^3 n}{2 \lambda^3}\right) \\
&\leq - \|\nabla Z_r(\theta_{t-1})\|^2 \frac{1}{n M_\ell} (-\psi'(0) Q_\infty r) + \|\nabla Z_r(\theta_{t-1})\| \left( \frac{M_\ell^2}{\lambda^2} + \frac{\ellbound^3 n}{2 \lambda^3}\right) \\
&\leq - \|\nabla Z_r(\theta_{t-1})\| \frac{\epsilon}{n M_\ell} (-\psi'(0) Q_\infty r) + \|\nabla Z_r(\theta_{t-1})\| \left( \frac{M_\ell^2}{\lambda^2} + \frac{\ellbound^3 n}{2 \lambda^3}\right) \\
&\leq - \|\nabla Z_r(\theta_{t-1})\| \frac{\epsilon}{n M_\ell} (-\psi'(0) Q_\infty r) \left( 1 -  \frac{M_\ell^3 n}{\lambda^2 \epsilon (-\psi'(0)) Q_\infty r} - \frac{\ellbound^4 n^2}{2 \lambda^3 \epsilon (-\psi'(0)) Q_\infty r} \right) \\
\end{align}

If the last term is positive then the whole expression is less than $0$ and the theorem is proved. A sufficient condition for this to be the case is that,

\begin{align}
\frac{M_\ell^3 n}{\lambda^2 \epsilon (-\psi'(0)) Q_\infty r} \leq \frac{1}{4},
\end{align}

and 

\begin{align}
\frac{\ellbound^4 n^2}{2 \lambda^3 \epsilon (-\psi'(0)) Q_\infty r} \leq \frac{1}{2},
\end{align}

which are both satisfied for 

\begin{align}
\lambda \geq \max \left( \sqrt{\frac{4 M_\ell^3 n}{\epsilon (-\psi'(0))Q_\infty r}} , \sqrt[3]{\frac{M_\ell^4 n^2}{\epsilon (-\psi'(0)) Q_\infty r}} \right).
\end{align}

\end{proof}


\subsubsection{Convergence}

\begin{thm}[\ref{thm:convergence}]
Assume that Assumptions \ref{assum:convex_loss}, \ref{assum:loss_regularity}, \ref{assum:smooth_bounded_q}, \ref{assum:local_linear_psi}, \ref{assum:smooth_score}, \ref{assum:bounded_support} hold
and that $\psi'(0) < 0$. If $\lambda = \Omega(n^{2/3})$ and $\nseed = o(n)$,
and if there exists some $C>0$ such that $\delta_{C,r,n} \rightarrow 0$ as $n \rightarrow \infty$, then for randomized uncertainty sampling parameters with fixed scale $r \leq m_\psi/M_q$, as $n \rightarrow \infty$,
\begin{align}
\nabla Z_r(\tilde{\theta}_n) \stackrel{P}{\rightarrow} 0.
\end{align}
\end{thm}
\begin{proof}

Assume that the parameter iterates are bounded by $C$ and the acceptance probability $a_r(\theta)$ is non-zero for all parameter iterates. This will occur with probability going to $1$ and so if we can show convergence in probability under this condition, then unconditional convergence in probability follows. The set of parameters that are bounded are a compact set and thus $\nabla Z_r(\theta)$ and $\nabla^2 Z_r(\theta)$ are bounded by some constant, call it $M_Z$.


From a Taylor expansion, for some $\theta'$ between $\theta_{t-1}$ and $\theta_t$,

\begin{align}
&Z_r(\theta_t) \\
&= Z_r(\theta_{t-1}) + \nabla Z_r(\theta_{t-1})^T (\theta_t - \theta_{t-1}) + (\theta_t - \theta_{t-1})^T \nabla^2 Z_r(\theta') (\theta_t - \theta_{t-1}) \\
&\leq Z_r(\theta_{t-1}) + \nabla Z_r(\theta_{t-1})^T (\theta_t - \theta_{t-1}) + \frac{M_\ell}{\lambda} M_Z \frac{M_\ell}{\lambda} \\
&\leq Z_r(\theta_{t-1}) - \nabla Z_r(\theta_{t-1})^T [\nabla^2 L_{t-1}(\theta_{t-1})]^{-1} \nabla \ell(z^{(t)}, \theta_{t-1}) + M_Z \left(\frac{M_\ell^2}{\lambda^2} + \frac{\ellbound^3 n}{2 \lambda^3} \right) + \frac{M_\ell^2 M_Z}{\lambda^2} \\
&\leq Z_r(\theta_{t-1}) - \nabla Z_r(\theta_{t-1})^T [\nabla^2 L_{t-1}(\theta_{t-1})]^{-1} \nabla \ell(z^{(t)}, \theta_{t-1}) + O(1/n)
\end{align}

Taking an expectation conditioned on $\theta_{t-1}$,

\begin{align}
\E[ Z_r(\theta_t) | \theta_{t-1}] &\leq Z_r(\theta_{t-1}) - \frac{-\psi'(0) Q_\infty r}{a_r(\theta_{t-1})} \nabla Z_r(\theta_{t-1})^T [\nabla^2 L_{t-1}(\theta_{t-1})]^{-1} \nabla Z(\theta_{t-1}) + O(1/n) \\
&\leq Z_r(\theta_{t-1}) - (-\psi'(0) Q_\infty r) \|\nabla Z_r(\theta_{t-1}) \|^2 \frac{1}{M_\ell (t-1)} + O(1/n)
\end{align}

Taking the expectation with respect to the randomness in the entire algorithm,

\begin{align}
\E[Z_r(\theta_t)] &\leq \E[Z_r(\theta_{t-1})] - \frac{-\psi'(0) Q_\infty r}{M_\ell} \E[ \|\nabla Z_r(\theta_{t-1}) \|^2 ] \frac{1}{t-1} + O(1/n).
\end{align}

Rearranging and summing over iterations $t \in [\nseed+1,n]$,

\begin{align}
\frac{-\psi'(0) Q_\infty r}{M_\ell} \sum_{t={\nseed}}^{n-1}  \frac{1}{t} \E[ \|\nabla Z_r(\theta_{t}) \|^2 ] &\leq \E[Z_r(\theta_{\nseed})] - \E[Z_r(\theta_n)] + O(1).
\end{align}

Since $Z_r$ is bounded between $0$ and $1$, then, for some constant $C$,

\begin{align}
\sum_{t={\nseed}}^{n-1}  \frac{1}{t} \E[ \|\nabla Z_r(\theta_{t}) \|^2 ] &\leq C \\
\frac{ \sum_{t={\nseed}}^{n-1}  \frac{1}{t} \E[ \|\nabla Z_r(\theta_{t}) \|^2 ] }{\sum_{t={\nseed}}^{n-1}  \frac{1}{t}} &\leq \frac{C}{ \sum_{t={\nseed}}^{n-1}  \frac{1}{t} } \\
\E[ \|\nabla Z_r(\tilde{\theta}_n)\|^2]   &\leq \frac{C}{\ln(\frac{n-1}{\nseed})}
\end{align}

Then, because $\nseed = o(n)$, the right side converges to $0$. So,

\begin{align}
\E[\|\nabla Z_r(\tilde{\theta}_n)\|^2] &\rightarrow 0 \\
\|\nabla Z_r(\tilde{\theta}_n)\|^2 &\stackrel{P}{\rightarrow} 0 \\
\|\nabla Z_r(\tilde{\theta}_n)\| &\stackrel{P}{\rightarrow} 0 \\
\nabla Z_r(\tilde{\theta}_n) &\stackrel{P}{\rightarrow} 0
\end{align}

\end{proof}

\subsection{Convergence of gradient}
\label{sec:convergence_of_gradient}

In this section, we wish to show that

\begin{align}
\lim_{r \rightarrow 0} \nabla Z_r(\theta) = \nabla Z(\theta)
\end{align}

Along with the results in the main paper, this result will enable us to say that uncertainty sampling is \emph{asymptotically} optimizing the zero-one loss.
First, we must define some concepts.
Each parameter vector $\theta$ defines a \emph{decision boundary}:
\begin{definition}[Decision Boundary]
\begin{align}
B_\theta &\eqdef \{x: S(x,\theta)=0\} .
\end{align}
\end{definition}

If $S(x,\theta)$ is differentiable with respect to $x$ and $\nabla_x
S(x,\theta) \neq 0$ for all $x \in B_\theta$, then by the implicit function
theorem, $B_\theta$ is a $(d-1)$-dimensional differentiable manifold and has
measure zero. 

\begin{proposition}
\label{prop:manifold}
If $S(x,\theta)$ is differentiable with respect to $x$ and $\nabla_x S(x,\theta) \neq 0$ throughout $B_\theta$, $B_\theta$ is an $(d-1)$-dimensional differentiable manifold and has measure zero.
\end{proposition}
\begin{proof}
For any point $b \in B_\theta$, since $\nabla_x S(x,\theta) \neq 0$, there is some direction where $\nabla_x S(x,\theta)$ is non-zero. By the implicit function theorem, this means that there is a differentiable mapping from a subset of $\mathbb{R}^{d-1}$ to a neighborhood of $b$ within $B_\theta$. Thus, $B_\theta$ is a $(d-1)$-dimensional differentiable manifold. Further, in $\mathbb{R}^d$, every open cover has a countable subcover. Thus, there is a countable family of local patches (with local differentiable charts). Since each local patch is a continuous mapping from a measure zero set $\mathbb{R}^{d-1}$, the local patches have measure zero. Since a countable union of measure zero sets has measure zero, $B_\theta$ has measure zero.
\end{proof}

When this
condition is satisfied, the decision boundary is well behaved, and $Z$ has nice properties. For these reasons, we will denote the set of parameters that meet this condition as the \emph{regular parameters}, $\validparams$:
\begin{definition}[Regular Parameters]
\begin{align}
\validparams \eqdef \{\theta: \forall x \in B_\theta, \nabla_x S(x,\theta) \neq 0\}.
\end{align}
\end{definition}
For logistic regression with identity features ($\phi(x)=x$),
$\nabla_x S(x,\theta) = \theta$, so 
the only point not in $\validparams$ is $\theta = 0$.
For logistic regression with quadratic features, $\theta \cdot \phi(x)=x^\top A x + b^\top x + c$
(the parameters are $A$, $b$, and $c$),
parameters where $A$ is non-singular and $c \neq \frac{1}{4} b^\top A^{-1} b$ are
in $\validparams$. Thus, the parameters not in $\validparams$ have measure zero.

The key quantity of interest will require an additional assumption. 
\begin{assumption}[Smooth PDF]
  \label{assum:continuous_pdf} $p^*(x,y)$ has a smooth (all derivatives exist) probability density function for both values of $y$.
\end{assumption}
Recall that the decision boundary $B_\theta$ has measure
zero for $\theta \in \validparams$. Assumption \ref{assum:continuous_pdf}
implies that there is zero probability mass on all decision boundaries corresponding to $\theta \in \validparams$ ($\BP(x \in B_\theta)=0$ for $x \sim p^*$).
This assumption and the ones from the main text ensure that the following quantity exists for $\theta \in \validparams$,

\begin{align}
\lim_{r \rightarrow 0} \frac{1}{2r} \int_{|S(x,\theta)|\leq r} (-y \nabla_\theta S(x,\theta)) dp^*(x,y)
\end{align}

This is a corollary of the following lemma,

\begin{lemma}
\label{lem:important_lemma}
Suppose $\theta \in \validparams$ and Assumption \ref{assum:smooth_score} holds. If $g(x)$ is smooth and has bounded support,

\begin{align}
F(s) = \int_{S(x,\theta)<s} g(x) dx
\end{align}

is smooth at $0$.
\end{lemma}
\begin{proof}

For this proof, we rely heavily on the arguments in \citet{hoveijn2007differentiability}

Since $g(x)$ has bounded support, for $\|x\| \geq M_x$, $g(x)=0$.
Intuitively, this means we can define a function that is equal to $S(x,\theta)$ for $\|x\|\leq M_x$ and is a small value $\|x\| \geq M_x$ and mollify to make it smooth. More precisely, let $S_{min} = \min(-2, \min_{\|x\| < 2 M_x} S(x,\theta))$. Define $f(x)$ to be equal to $S(x,\theta)$ inside a ball of radius $2 M_x$ and equal to $S_{min}$ outside. Then mollify the function between balls of radius $M_x$ and $2M_x$. If we shift the function by $S_{min}$, the function is smooth, always positive, and vanishes at infinity. Thus, it satisfies the Shifted class C functions of Definition 2 of \citet{hoveijn2007differentiability}.

Then, we can examine the function 

\begin{align}
G(s) = \int_{-1<f(x)<s} g(x) dx,
\end{align}

which will have the same derivatives (if they exist) as $F(s)$ around $0$. Note that $S_{min} \leq -2 < -1$, so the integration between the level sets is well-defined.

$0$ is a regular value because $\theta \in \validparams$. Further,
we don't need the non-degeneracy conditions of \citet{hoveijn2007differentiability} because $\nabla_x S(x,\theta)$ is continuous (Assumption \ref{assum:smooth_score}) on a compact set (the support of $g(x)$) and thus is bounded below. And thus, a neighborhood around $0$ are regular values.

We can use the flow box and diffeomorphism argument from \citet{hoveijn2007differentiability} to express the volume function as an integral with $h$ as the upper limit (see Proposition 7 of \citet{hoveijn2007differentiability}). While \citet{hoveijn2007differentiability} uses $1$ as the integrand, the same argument holds for $g(x)$ as the integrand, and we recover that since $g(x)$ is smooth, the integral is smooth.

\end{proof}

\begin{proposition}
If Assumptions \ref{assum:smooth_score}, \ref{assum:bounded_support}, and \ref{assum:continuous_pdf} hold, then for $\theta \in \validparams$,

\begin{align}
\lim_{r \rightarrow 0} \frac{1}{2r} \int_{|S(x,\theta)|\leq r} (-y \nabla_\theta S(x,\theta)) dp^*(x,y)
\end{align}

exists.

\end{proposition}
\begin{proof}
Let $g_1(x) = \nabla_\theta S(x,\theta) p^*(x,y=-1)$ and $g_2(x) = \nabla_\theta S(x,\theta) p^*(x,y=1)$. Let the corresponding integrals according to Lemma \ref{lem:important_lemma} be $F_1(s)$ and $F_2(s)$.

\begin{align}
& \lim_{r \rightarrow 0} \frac{1}{2r} \int_{|S(x,\theta)|\leq r} (-y \nabla_\theta S(x,\theta)) dp^*(x,y) \\
&= \lim_{r \rightarrow 0} \frac{1}{2r} \int_{|S(x,\theta)|\leq r} \nabla_\theta S(x,\theta) dp^*(x,y=-1) - \lim_{r \rightarrow 0} \frac{1}{2r} \int_{|S(x,\theta)|\leq r} \nabla_\theta S(x,\theta) dp^*(x,y=1) \\
&= F_1'(s) - F_2'(s)
\end{align}
\end{proof}

Now it remains to show that both $\lim_{r \rightarrow 0} \nabla Z_r(\theta)$ and $\nabla Z(\theta)$ are equal to that quantity for $\theta \in \validparams$.

First, to show the result about $\nabla Z_r$, we will assume $q$ is smooth.
\begin{assumption}
\label{assum:q_smooth}
$q$ is smooth, that is, all derivatives exist.
\end{assumption}

\begin{proposition}
If Assumptions \ref{assum:smooth_bounded_q}, \ref{assum:smooth_score}, \ref{assum:bounded_support}, \ref{assum:continuous_pdf}, and \ref{assum:q_smooth} hold, then for $\theta \in \validparams$,
\begin{align}
\lim_{r \rightarrow 0} \nabla Z_r(\theta) = \lim_{r \rightarrow 0} \frac{1}{2r} \int_{|S(x,\theta)|\leq r} (-y \nabla_\theta S(x,\theta)) dp^*(x,y)
\end{align}
\end{proposition}
\begin{proof}
Recall that $q$ is an even function with bounded support. Thus, let $M_q$ be a number such that $|x| \geq M_q \implies q(x)=0$. Then,

\begin{align}
q(x) &= q(|x|) \\
&= \int_{M_q}^{|x|} q'(u) du \\
&= - \int_{|x|}^{M_q} q'(u) du \\
&= - \int_{0}^{M_q} \textbf{1}[u \geq |x|] q'(u) du
\end{align}

From the proof of Theorem \ref{thm:key_observation}

\begin{align}
\nabla Z_r(\theta) &= \frac{1}{Q_\infty r} \int q(S(x,\theta)/r) (-y \nabla_\theta S(x,\theta)) dp^*(x,y) \\
&= - \frac{1}{Q_\infty r} \int  \int_0^{M_q} \textbf{1} \left[u \geq \left|\frac{S(x,\theta)}{r}\right| \right] q'(u) (-y \nabla_\theta S(x,\theta)) du dp^*(x,y) \\
&= - \frac{1}{Q_\infty r} \int_0^{M_q}  \int  \textbf{1}[|S(x,\theta)| \leq ru] q'(u) (-y \nabla_\theta S(x,\theta)) dp^*(x,y) du \\
&= - \frac{1}{Q_\infty} \int_0^{M_q} 2 u q'(u) \frac{1}{2ru} \int_{|S(x,\theta)| \leq ru} (-y \nabla_\theta S(x,\theta)) dp^*(x,y) du 
\end{align}

We can can the limit as $r \rightarrow 0$ of both sides. Further, we can use the dominated convergence theorem to bring the limit inside the outer integral.

\begin{align}
\lim_{r \rightarrow 0} \nabla Z_r(\theta) &= - \frac{1}{Q_\infty} \int_0^{M_q} 2 u q'(u) \lim_{r \rightarrow 0} \frac{1}{2ru} \int_{|S(x,\theta)| \leq ru} (-y \nabla_\theta S(x,\theta)) dp^*(x,y) du \\
&= - \frac{1}{Q_\infty} \int_0^{M_q} 2 u q'(u) du \lim_{r \rightarrow 0} \frac{1}{2r} \int_{|S(x,\theta)| \leq r} (-y \nabla_\theta S(x,\theta)) dp^*(x,y) \\
\end{align}

Now the only thing left to show is that the integral with respect to $u$ is $1$ and the result will follow. We can use integration by parts:

\begin{align}
- \frac{1}{Q_\infty} \int_0^{M_q} 2 u q'(u) du &= - \frac{2}{Q_\infty} \left( [q(u) u ]_0^{M_q} - \int_0^{M_q} q(u)  du \right) \\
&= - \frac{2}{Q_\infty} \left(0 \cdot M_q - q(0) \cdot 0 - \frac{Q_\infty}{2} \right)\\
&= 1 \\
\end{align}
\end{proof}

\begin{proposition}
If Assumptions \ref{assum:smooth_score}, \ref{assum:bounded_support}, and \ref{assum:continuous_pdf} hold, then for $\theta \in \validparams$,
\begin{align}
\nabla Z(\theta) = \lim_{r \rightarrow 0} \frac{1}{2r} \int_{|S(x,\theta)| \leq r} (-y \nabla_\theta S(x,\theta)) dp^*(x,y).
\end{align}
\end{proposition}
\begin{proof}
Noting that the derivative of the Heaviside step function is the Dirac delta function and that the Dirac delta function is even,

\begin{align}
Z(\theta) &= \int H(-yS(x,\theta)) dp^*(x,y) \\
\nabla Z(\theta) &= \int \delta(-yS(x,\theta)) (-y \nabla_\theta S(x,\theta)) dp^*(x,y) \\
&= \int \delta(S(x,\theta)) (-y \nabla_\theta S(x,\theta)) dp^*(x,y)
\end{align}

\begin{align}
& \lim_{r \rightarrow 0} \frac{1}{2r} \int_{|S(x,\theta)| \leq r} (-y \nabla_\theta S(x,\theta)) dp^*(x,y) \\
&= \lim_{r \rightarrow 0} \int \frac{\textbf{1}[-r \leq S(x,\theta) \leq r]}{2r} (-y \nabla_\theta S(x,\theta)) dp^*(x,y) \\
&= \int \delta(S(x,\theta)) (-y \nabla_\theta S(x,\theta)) dp^*(x,y) \\
&= \nabla Z(\theta)
\end{align}
\end{proof}

Combining the previous propositions, we find that,

\begin{theorem}
If Assumptions \ref{assum:smooth_bounded_q}, \ref{assum:smooth_score}, \ref{assum:bounded_support}, \ref{assum:continuous_pdf}, and \ref{assum:q_smooth} hold, then for $\theta \in \validparams$,
\begin{align}
\lim_{r \rightarrow 0} \nabla Z_r(\theta) = \nabla Z(\theta).
\end{align}
\end{theorem}

\end{document}